# Automatic Summarization of Natural Language
## Literature Review and Synthesis

Marc Everett Johnson[1]

*Minneapolis, Minnesota, USA*

`me@marc.mn`

*Abstract*— Automatic summarization of natural language is a current topic in computer science research and industry, studied for decades because of its usefulness across multiple domains. For example, summarization is necessary to create reviews such as this one. Research and applications have achieved some success in *extractive* summarization (where key sentences are curated), however, *abstractive* summarization (synthesis and re-stating) is a hard problem and generally unsolved in computer science. This literature review contrasts historical progress up through current state of the art, comparing dimensions such as: *extractive vs. abstractive, supervised vs. unsupervised, NLP (Natural Language Processing) vs Knowledge-based, deep learning vs algorithms, structured vs. unstructured sources, and measurement metrics such as Rouge and BLEU*. Multiple dimensions are contrasted since current research uses combinations of approaches as seen in the review matrix. Throughout this summary, synthesis and critique is provided. This review concludes with insights for *improved abstractive summarization measurement*, with surprising implications for *detecting understanding and comprehension in general*.

## I. INTRODUCTION

What is summarization? When asked to summarize, many people intuitively understand the requirement to shorten content while preserving important information. One might think of jpeg digital picture compression, which reduces the file size while preserving fidelity. Similarly, some describe forms of summarization as sentence or word compression, which gives us insight into one way to perform summarization. Abstractive summarization also synthesizes new words and sentences, while preserving underlying meaning -- a much harder problem -- which this review will also discuss.

The focus of this review is natural language summarization, like is performed by literature digests, or akin to writing this paper summarizing and synthesizing other literature.

## II. LITERATURE REVIEW

For this literature review, multiple journal articles are analyzed leading to future research questions and implications.

The matrix on the following page visually illustrates recent trends and contrasts across topics of research in automatic natural language summarization. Colors delineate trends per column. Most of columns below each have a corresponding section in the following pages that discusses the results in detail. In addition to the contrasts and trends below, new avenues and insights are discussed in the pages following.

TABLE I

AUTOMATIC SUMMARIZATION OF NATURAL LANGUAGE - COMPARISON MATRIX

| Reference | Extract vs Abstract | Supervised vs Unsupervised | Deep vs Algo | Structured vs Unstruct'd Source | Single vs Multi Source | Metric | Rouge-2 Score | Data Set | Summary Units |
|---|---|---|---|---|---|---|---|---|---|
| **Ganesan et al, 2010 (IBM)** | Abstract | Unsupervised | Algo Graph | Unstructured | Multi | Rouge | 0.0998 | Opinions | Sentence |
| **Yao et al, 2015 (NNSF)** | Extract | Unsupervised | Algo ADMM | Unstructured | Multi | Rouge | 0.1245 | DUC 06/07 | Document |
| **Rush et al, 2015 (Harvard)** | Abstract | Supervised | Deep NNLM | Structured | Single | Rouge | 0.1265 | News / Gigaword | Sentence |
| **Hermann et al, 2015 (Google)** | Abstract | Supervised | Deep LSTM | Structured | Single | Word Dist | n/a | News / CNN | Sentence |
| **Chopra et al, 2016 (FB)** | Abstract | Supervised | Deep RNN | Structured | Single | Rouge | 0.1597 | News / Gigaword | Sentence |
| **Banarjee et al, 2016 (NSF)** | Abstract | Unsupervised | Algo ILP | Unstructured | Multi | Rouge | 0.1199 | DUC 04/05 | Document |
| **Nallapati et al, 2016 (IBM)** | Extract | Unsupervised (partial) | Deep RNN | Structured | Single | Rouge | 0.162 | News / CNN | Sentence |
| **Yousefi-Azar et al, 2017** | Extract | Unsupervised | Deep AE | Unstructured | Single | Rouge | 0.1647 | Emails / SKE + BC3 | Document (email) |
| **Zhou et al, 2017 (Microsoft)** | Abstract | Supervised | Deep GRU | Structured | Single | Bleu | 13.29 (Bleu) | Q & A SQuAD | Document |
| **Dohare et al, 2017 (Microsoft)** | Abstract | Supervised | Deep AMR | Structured | Single | Rouge | 0.1575 | News / CNN | Document |
| **Paulus et al, 2017 (Salesforce)** | Abstract | Supervised | Deep LSTM | Structured | Single | Rouge | 0.1582 | News / CNN | Word |

## A. Extractive vs. Abstractive

A primary contrast between summary approaches is *extractive* vs *abstractive*.

*Extractive* summarizations is the easiest to understand and perform, because extraction is a subset of the input text -- for example removing sentences and words. Academically you might call this "sentence ranking" (Yousefi-Azar & Hamey, 2016) or "Greedy, Constraint, Graph Based" (Nallapati, Zhai, & Zho, 2017). A challenging part of extractive summarization is to identify less important sentences and words in order to remove them.

*Abstractive* summarization is a hard problem, because new sentences are created and synthesized. For example, guidelines for a school writing assignment may state "the paper should not be a mere recitation of information… rather it should be a synthesis of ideas… showing you have understood and assimilated the subject matter". This neatly illustrates the difference between extractive ("recitation") and abstractive ("synthesis").

*Abstraction* is harder than *extraction* for humans and computers. Abstractive summarization can require "Prior knowledge, natural language processing and understanding" (Ganesan, Zhai, & Han 2010). In prior decades' research abstractive solutions were rarer, but with advances in deep learning these systems are more commonplace. The comparison matrix above (Table 1) shows this trend.

*B. Supervised vs Unsupervised Learning*

In machine learning, *supervised* learning uses datasets to train, whereas *unsupervised* learning does not (or uses latent features). Supervised and unsupervised approaches can be categorized into the following groups: "latent topic models" for unsupervised techniques, and "classification and regression" as the supervised techniques (Yousefi-Azar et al., 2016).

A supporting insight is borne out in the comparison matrix. Extractive summarization are almost always achieved *unsupervised* -- which can be seen in the studies shown in Table 1. This is likely because these solutions are largely traditionally algorithmic or NLP-based. *Abstractive* techniques often (but not always) use supervised learning, since custom rubrics are still required for abstraction. Some groups have achieved unsupervised learning in abstractive training (Yousefi-Azar et al., 2017) via "deep auto-encoder (AE) to learn features rather than manually engineering them". This is a significant efficiency and area for future progress.

*C. NLP vs knowledge-based*

*Natural language processing (NLP)* research dates back to 1950's. NLP uses sentence grammars, ontologies, language models, parse trees and similar methods to analyze and process language. NLG (natural language generation) does the reverse, generating natural language from machine representation. NLP/NLG based summaries are sometimes called "semantic-based or ontology based" (Allahyari et al., 2017) to contrast with *"knowledge-based"*.

Prior to deep learning (which are often *knowledge-based*) systems, NLP and ontology-based solutions were the most popular avenue to attempt abstractive transformations. For example, sentences might be merged via mechanistic conjunctive language rules. This type of abstraction is mostly grammatical, and may not provide synthesis of the documents ideas. Consider this "abstraction-light".

Some researchers combine NLP with deep learning where they "encode linguistic information" including POS (parts of speech) and NER (named entity recognition) tags as the lexical features as part of the neural encoder-decoder neural network (Zhou, Yang, Wei, Tan, & Bao, 2017). I agree with Allahyari et al (2017) who observes that "a step towards building more accurate summarization systems is to combine summarization techniques with *knowledge bases* and semantic-based or *ontology-based* summarizers." A trend that can be seen in the comparison matrix is a pivot away from NLP and more towards Deep Learning.

*D. Deep Learning vs Algorithms*

In early research, many solutions used *algorithmic*, NLP and NLG techniques as touched upon in the previous section.

*Deep learning* models have historically proven effective for machine translation and speech recognition. Now summarization is treated as a training and classification problem as well. Google, Facebook, IBM, Microsoft and other companies are developing successful models based on Recurrent Neural Network (RNN), convolutional neural network (CNN), as well as LSTM, NNLM, AMR, GRU, AE network models documented in the matrix (Table 1).

Not only does abstractive summarization benefit from deep learning; some extractive techniques now use automated training for determining sentence ranking "eliminating the need for sentence-level extractive labels" (Nallapati et al., 2016).

Purely algorithmic approaches are still being developed. For example ADMM (alternating direction method of multipliers, a general algorithm that solves convex optimization problems) delivers effective extractive summarization (Yao, Wan, & Xiao, 2015). Areas of overlap between NLP, Deep Learning and traditional algorithms are a fruitful avenue of research.

*E. Structured vs Unstructured Sources*

Another facet of comparison for summarization problems are *structured* vs *unstructured* sources. For example, the format of a "Scrum" meeting may be structured "What did I do yesterday, what will I do today, and what am I blocked on?" A summary needs to parse and honor this format. An insight from the review is that although abstractive work is increasing, it is resting in *structured sources*, such as news and meetings. Facebook used the relation between news stories and headlines to train an abstractive model to predict headlines (Rush, Chopra, & Watson, 2015).

An early but powerful approach is Opinosis (Ganesan et al., 2010) which is a "graph-based summarization framework that generates abstractive summaries of highly redundant opinions." What was groundbreaking is its simple ability to create abstractions algorithmically. It assumes no domain knowledge, but works best on opinions (due to their repetitiveness).

Both of these examples illustrate an opportunity area or research for summarization to adapt dynamically to domain and structure. The comparison matrix reveals that many of the training sources are "CNN" news, which may not make for robust generalization.

*F. Metrics*

Fitness of summarization needs to be measured. Measurement in itself can be challenging if the summarization involves abstractions. A metric ideally works for different types of summaries or languages.

The most popular measure is called "Rouge" which measures recall and how much the words appear in the reference. Another method is called "Bleu" which measures precision (how words match the reference summaries). The Google DeepMind team developed their own benchmark to measure word distance (Hermann et al., 2015).

Even with a metric like Rouge, a reference dataset is required to compare summary fitness, akin to a grading key. If the reference summaries are manually generated, this can impart subjectivity (Ferreira, 2013). Several approaches use cloud services like Amazon's Mechanical Turk to generate the reference summary. Interestingly, some new approaches auto-generate references (Yousefi-Azar et al., 2017) by attempting to learn features.

The comparison matrix includes Rouge scores where available. A general trend shows higher scores for extractive approaches, which makes sense since extraction is more mature and better understood. An improving metric trend for abstraction is shown over time as well. Note that studies with similar methodologies and data sets are most comparable.

A potential issue is "gaming" where focus on discrete metrics "increases score without an actual increase in readability or relevance." (Paulus, Xiong, & Socher, 2017). Several articles made comments relating to this: a single metric can be detrimental to the model quality. I feel this measurement shortcoming is a fruitful area of research.

This concludes the review of prior work, and leads us to future directions for research.

## III. Future Research Directions

As discussed in the prior section *Rouge* and *Bleu* (and other metrics) are measuring tools, but require a reference dataset to compare summary fitness, akin to a grading key (Ferreira, 2013). "Gaming" occurs when an algorithm improves score without improving readability. Several articles made comments relating to this: a single metric can be detrimental to the model quality.

This leads to a possible research question: *"How can an abstractive summarization **measurement** rubric be automatically generated and scored, without susceptibility to gaming or over-training?"*

Why is is the *automatic* measurement question important? Without automation, it is difficult to know if abstractive summarization or synthesis is correct when using *manual* processes. Reference summaries are manually generated today which is subjective as well as not scalable. Specifically, automatic measurement is essential for objective feature learning, training and model quality en route to summarization and intelligence.

Automatic training of deep learning models (for Automatic Summarization of Natural Language) could follow these future directions:

1. Auto-generate references (Yousefi-Azar et al., 2017) by attempting to learn features.
2. The Google DeepMind team developed their own benchmark to measure word distance (Hermann et al., 2015).
3. Opinosis algorithm (Ganesan et al., 2010) is a studied but limited algorithmic summarization method usable as a reference for newer deep learning methods (e.g. use known algorithms to grade others).
4. Narrow measurement rubric from general natural language abstractive summarization to specific extractive summarization metric in limited domains -- for example for SCRUM meetings (Rush, Chopra, & Watson, 2015), source code summarization (Moreno & Marcus, 2017), question generation (Zhou et al., 2017) or improving search queries and results (Hermann et al., 2015).

When the automatic training and measurement problem is solved, this may provides help with measurement and *detection of machine comprehension and intelligence itself.* This is because a machine intelligence test called "I-Athalon" (Adams, 2016) proposes that summarizing is a key test criteria towards proving machine understanding (Marcus, 2017). Hermann et al (2015) goes so far to characterize work on abstractive summarization (for Google DeepMind) as "machine comprehension." (!)

As such, another research question may arise: *"How can abstractive summarization and its effective measurement provide key criteria for machine intelligence tests, towards proving machine understanding and comprehension (ala improved classic Turing Test)?"*

Addressing this second question is interesting because the Turing test is considered a classic threshold AI will pass en route to human-level intelligence. Examining this could have cross disciplinary application in cognitive and behavioral sciences towards automatic measurement of human and animal intelligence as well.

This is a leading edge area, so there isn't much prior research, but some avenues include:

1. Starting with the results of the measurement research question, utilize its rubrics as an automated component towards assessing machine comprehension (per insights of Adams, Marcus and Hermann).
2. Measure machine comprehension by drawing upon cross-disciplinary approaches employed for human and animal understanding. This collaboration would contribute back to other fields, since automated measurement of comprehension is useful generally.

## IV. Closing Thoughts

This literature review contrasted and synthesized recent developments in automatic summarization. Advances in abstractive summarizers and deep learning systems are observed. Extractive techniques continue to achieve top fitness scores, while a progressing metric trend for abstraction is closing the gap.

Opportunity areas include improving unsupervised learning for diverse sources, blending NLP vs knowledge based insights, and improving measurement metrics.

Research data sets commonly include news and meetings. Additional data and problem domains include question generation (Zhou et al., 2017), source code summarization (Moreno & Marcus, 2017) and improving search queries and results (Hermann et al., 2015) to list a few.

Abstractive summarization measurement rubrics that are *automatically* generated and scored, without susceptibility to gaming or over-training will be a general accelerator for the field.

An important (and perhaps surprising) implication of improved abstraction measurement is the resultant contribution to automated intelligence tests, for *detecting understanding and comprehension in general.*

REFERENCES

[1] Adams, S. S., Banavar, G., & Campbell, M. (2016, March). I-athlon: toward a multidimensional Turing Test. AI Magazine, 37(1), 78-85.
[2] Allahyari, M., Pouriyeh, S., Assefi, M., Safaei, S., Trippe, E. D., Gutierrez, J. B., & Kochut, K. (2017, July). Text summarization techniques: A brief survey. arXiv preprint arXiv:1707.02268.
[3] Banerjee, S., Mitra, P., & Sugiyama, K. (2016, September). Multi-Document Abstractive Summarization Using ILP Based Multi-Sentence Compression. In IJCAI. 1208-1214.
[4] Chopra, S., Auli, M., & Rush, A. M. (2016, June). Abstractive sentence summarization with attentive recurrent neural networks. In Proceedings of the 2016 Conference of the North American Chapter of the Association for Computational Linguistics: Human Language Technologies. 93-98.
[5] Dohare, S., & Karnick, H. (2017, July). Text Summarization using Abstract Meaning Representation. arXiv preprint arXiv:1706.01678.
[6] Ferreira, R., de Souza Cabral, L., Lins, R. D., e Silva, G. P., Freitas, F., Cavalcanti, G. D., Favaro, L. (2013, March). Assessing sentence scoring techniques for extractive text summarization. Expert systems with applications, 40(14), 5755-5764.
[7] Ganesan, K., Zhai, C., & Han, J. (2010, August). Opinosis: a graph-based approach to abstractive summarization of highly redundant opinions. In Proc. of 23rd international conf. on computational linguistics. Association for Computational Linguistics. 340-348.
[8] Hermann, K. M., Kocisky, T., Grefenstette, E., Espeholt, L., Kay, W., Suleyman, M., & Blunsom, P. (2015, December). Teaching machines to read and comprehend. Advances in Neural Information Processing Systems, 1693-1701.
[9] Marcus G. (2017, March). The Search for a New Test of Artificial Intelligence. In Scientific American Magazine, 316(3), 59-63.
[10] Moreno, L., & Marcus, A. (2017, May). Automatic software summarization: the state of the art. In Proceedings of the 39th International Conference on Software Engineering Companion. IEEE Press. 511-512.
[11] Nallapati, R., Zhai, F., & Zhou, B. (2016, November). SummaRuNNer: A recurrent neural network based sequence model for extractive summarization of documents. In Proceedings of AAAI-17. 3075-3081.
[12] Paulus, R., Xiong, C., & Socher, R. (2017, November). A Deep Reinforced Model for Abstractive Summarization. arXiv preprint arXiv:1705.04304.
[13] Rush, A. M., Chopra, S., & Weston, J. (2015, September). A neural attention model for abstractive sentence summarization. arXiv preprint arXiv:1509.00685.
[14] Yao, J. G., Wan, X., & Xiao, J. (2015, June). Compressive Document Summarization via Sparse Optimization. In IJCAI. 1376-1382.
[15] Yousefi-Azar, M., & Hamey, L. (2017, February). Text summarization using unsupervised deep learning. Expert Systems with Applications, 68, 93-105.
[16] Zhou, Q., Yang, N., Wei, F., Tan, C., Bao, H., & Zhou, M. (2017, April). Neural Question Generation from Text: A Preliminary Study. arXiv preprint arXiv:1704.01792.